\documentclass[letterpaper, 10 pt, conference]{ieeeconf}  

\IEEEoverridecommandlockouts                              

\usepackage{makecell}
\usepackage{cite}
\usepackage{amsmath,amssymb,amsfonts}
\usepackage{algorithmic}
\usepackage{graphicx}
\usepackage{textcomp}
\usepackage{xcolor}
\usepackage{academicons}
\usepackage{siunitx}
\usepackage{booktabs}
\usepackage{multirow}
\usepackage{svg}
\usepackage{amsmath}
\usepackage{wasysym}
\usepackage{caption}
\usepackage[normalem]{ulem}
\usepackage[linesnumbered,ruled,vlined]{algorithm2e}
\newbox{\myorcidaffilbox}
\sbox{\myorcidaffilbox}{\large\includegraphics[height=1.25ex]{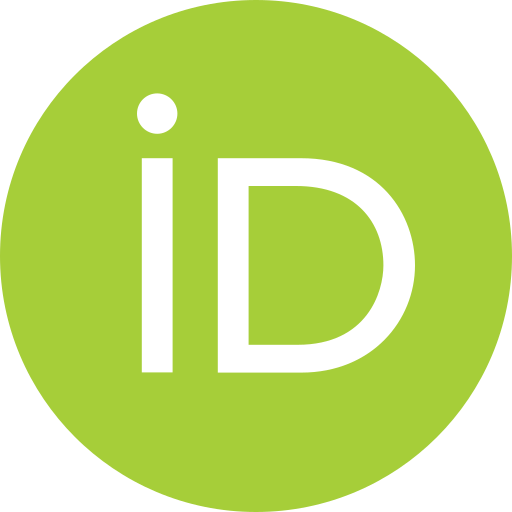}}
\newcommand{\orcidaffil}[1]{%
  \href{https://orcid.org/#1}{\usebox{\myorcidaffilbox}}}

\newcommand{\comment}[1]{\ignorespaces}

\makeatletter
\let\NAT@parse\undefined
\makeatother
\usepackage{hyperref}

\begin{document}
\title{\LARGE \bf
Autonomously Unweaving Multiple Cables Using Visual Feedback
}


\author{Tina Tian$^{1}$,
        Xinyu Wang$^{2}$, 
        Andrew L. Orekhov$^{1}$,
        Fujun Ruan$^{1}$,
        Lu Li$^{1}$,
        Oliver Kroemer$^{2}$,
        Howie Choset$^{1}$
\thanks{$^{1}$\textit{Robotics Institute, Carnegie Mellon University, 5000 Forbes Ave, Pittsburgh, PA 15213, United States} }  
}

\twocolumn[{
\renewcommand\twocolumn[1][]{#1}%
\maketitle
\begin{center}
    \centering
    \maketitle
    \includegraphics[width=\textwidth]{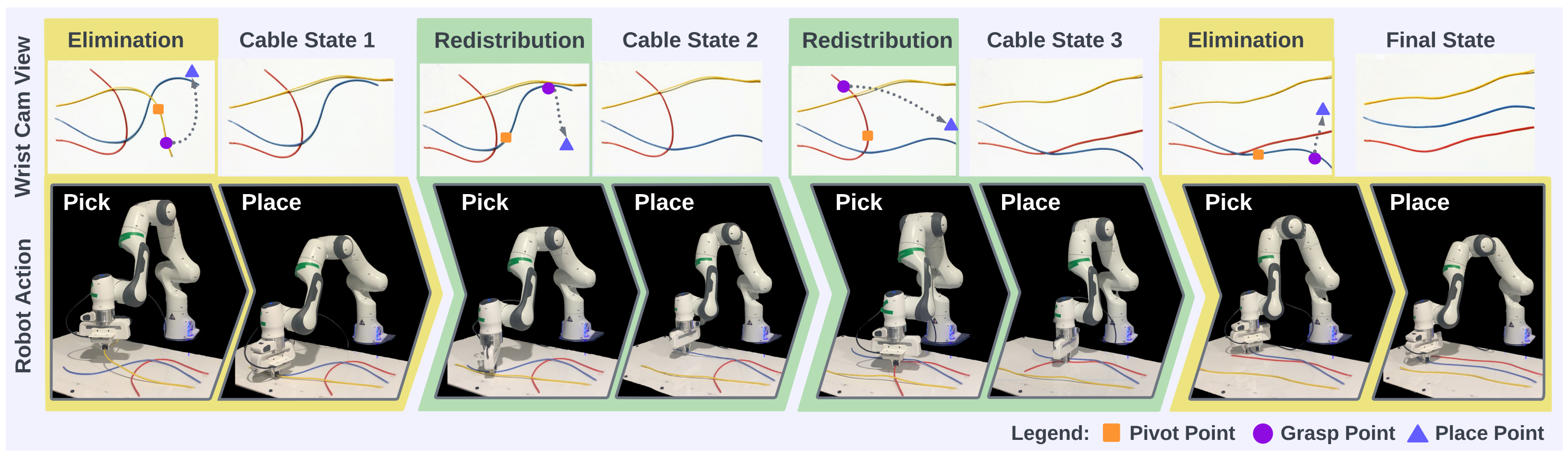}
    \vspace{-10 pt}
    \captionsetup{type=figure}
    \captionof{figure}{Our system autonomously unweaves multiple cables using visual feedback and a combination of two action primitives, \textit{elimination} for crossing elimination and \textit{redistribution} for rearranging and even distribution of the cables, in an iterative manner until the cables are unwoven and no crossing is present.}
    \label{fig:cover}
\end{center}

}]

{
  \renewcommand{\thefootnote}%
    {\fnsymbol{footnote}}
  \footnotetext{$^{1}$ \textit{Biorobotics Lab, Carnegie Mellon University, 5000 Forbes Ave, Pittsburgh, PA 15213, United States}}
  \footnotetext{$^{2}$ \textit{Intelligent Autonomous Manipulation Lab, Carnegie Mellon University, 5000 Forbes Ave, Pittsburgh, PA 15213, United States}}
}
\begin{abstract}
Many cable management tasks involve separating out the different cables and removing tangles. Automating this task is challenging because cables are deformable and can have combinations of knots and multiple interwoven segments. Prior works have focused on untying knots in one cable, which is one subtask of cable management. However, in this paper, we focus on a different subtask called \emph{multi-cable unweaving}, which refers to removing the intersections among multiple interwoven cables to separate them and facilitate further manipulation. We propose a method that utilizes visual feedback to unweave a bundle of loosely entangled cables. We formulate cable unweaving as a pick-and-place problem, where the grasp position is selected from discrete nodes in a graph-based cable state representation. Our cable state representation encodes both topological and geometric information about the cables from the visual image. To predict future cable states and identify valid actions, we present a novel state transition model that takes into account the straightening and bending of cables during manipulation. Using this state transition model, we select between two high-level action primitives and calculate predicted immediate costs to optimize the lower-level actions. We experimentally demonstrate that iterating the above perception-planning-action process enables unweaving electric cables and shoelaces with an 84\% success rate on average.
\end{abstract}



\section{Introduction}
Cable management involves the process of untangling and organizing bundles of cables in a systematic manner. It has a wide range of applications in the automotive industry, IT infrastructure, and hospital \cite{industry,wireharness}. Automating cable management is challenging because cables and ropes, as part of a class of objects called deformable linear objects (DLOs), have infinite dimensionality and intricate deformation behaviors, and they can present complicated knots and intertwinements. Prior work on cable management has primarily focused on the subtask of untying knots on a single cable \cite{viswanath2021singledense,she2020rss,viswanath2022long}. By contrast, our work focuses on the underexplored subtask of unweaving multiple cables, with the goal of eliminating crossings between cables to organize the cables. This capability provides a first step in the cable management process, after which prior methods of untying knots in a single cable could then be applied.

In this paper, we propose an approach for unweaving a collection of intersecting cables to eliminate all crossings between them. We assume that the cables 1) are visually distinct (we use different known colors), 2) lie loosely on a 2D plane, 3) are all fixed to the environment on one end (e.g. electrical or data cables with one end inserted into a connector), and 4) only contain bends and crossings and do not contain self-loops or knots. Our approach utilizes a single robot arm to perform the cable unweaving task.

\begin{figure}[ht]
    \centerline{
    \includegraphics[width=\linewidth]{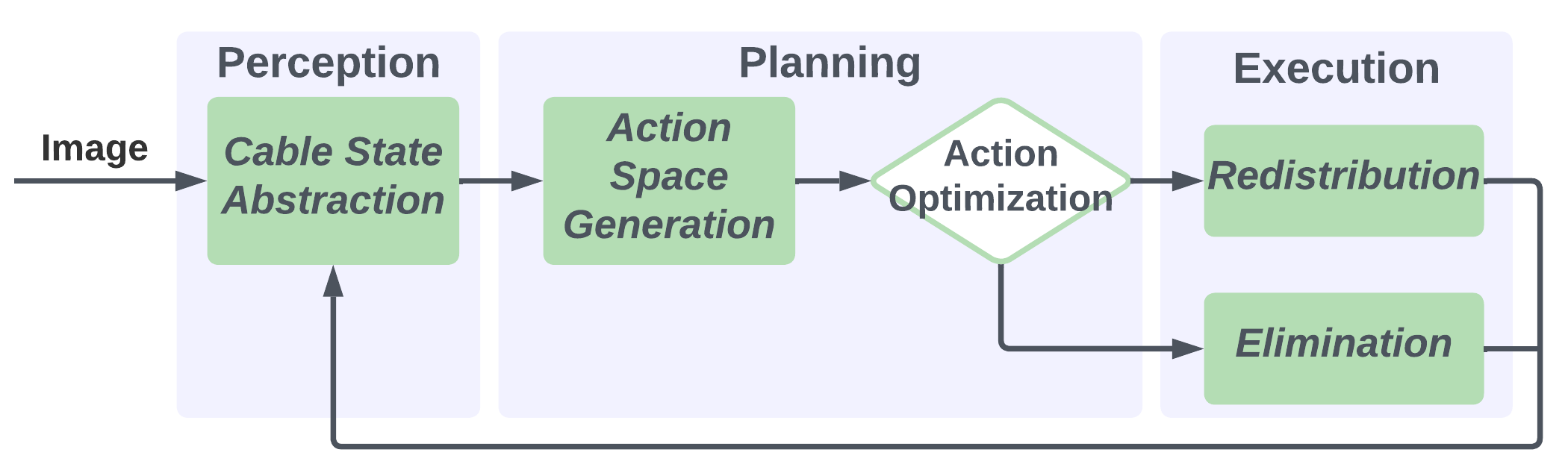}
    }
    \caption{Our pipeline constructs a cable state graph representation from an RGBD image obtained from wrist camera and generates actions that iteratively unweave mutliples cables.}
    \label{fig:flowchart}
\end{figure}

We address the multi-cable unweaving problem by leveraging a graph-based cable state representation with discrete nodes abstracted from visual image data, a state transition model, and an optimization-based planner (Fig. \ref{fig:cover}). After acquiring RGBD data from an Intel RealSense D435 camera that is rigidly attached to the arm's end-effector, we abstract the geometric and occlusion relationships of the cables in the image into a set of cable graphs using sliding-window-based visual tracing. We then identify the valid actions by checking the next state of each action using our proposed transition model. To plan a pick-and-place action, we first choose one of two action primitives, crossing elimination or redistribution, depending on the distribution of the cables. Subsequently, we select the optimal action parameters by minimizing the immediate predicted cost. After executing the action, we return to the sensing and perception step to acquire the new cable state and repeat this process until no crossing is detected. The flowchart of our multi-cable unweaving method is shown in Fig. \ref{fig:flowchart}. 

We evaluate our method on cable unweaving tasks with different numbers of cables, initial cable layouts, and materials. Experimental results suggest that our method achieves a cable state identification success rate of 99\% and a successful unweaving rate of 84\%.

Our contributions are summarized as follows:
\begin{itemize}
\item A graph-based cable state representation that encodes both topological and geometric information of multiple interwoven cables.
\item A deterministic state transition model that utilizes segment-wise graph straightening.
\item An optimization-based action selection method using two action primitives, which we call \textit{elimination} and \textit{redistribution}, for iterative cable unweaving.
\end{itemize}
\section{Related Work}
One of the fundamental problems in cable manipulation is defining a suitable representation of the cable state. Like many works on visuomotor skill learning, Nair et al. \cite{nair2017combining} learn cable embeddings into a latent representation and train pixel-level manipulation policies. However, this method does not fully utilize DLO's unique topological features. Schulman et al. \cite{schulman2013tracking} represent the cable using the geometric coordinates of each node, with no explicit encoding of the cable occlusion relationship. Other prior works have taken inspiration from Knot Theory \cite{manturov2020knottheory}, which provides means to capture and analyze the topological states of a DLO \cite{saha2006dloplanning, lui2013tangled, viswanath2021singledense}. To map sensor data into cable state representations, there are two popular methods: 3D cable model fitting \cite{lui2013tangled} and 2D visual tracing \cite{viswanath2022long}. We use a 2D visual tracing method similar to \cite{viswanath2022long}, using a sliding-window-based discretization and online keypoint identification with one end of the cables fixed.

The second challenge in cable manipulation is predicting the future cable states through a state transition model, which is crucial for planning manipulation actions. In \cite{aprilTag}, an interaction network is used to learn DLO's full 3D dynamics by tracking Aruco markers with Microsoft Kinect. Wang et al. \cite{wang2022offlineOnline} propose a combination of offline learning using a graph neural network in simulation to acquire the dynamics and an online linear residual model to bridge the sim-to-real gap. Another type of method for future state prediction leverages analytical transition models. In \cite{lui2013tangled}, a hand-crafted transition model is proposed based simply on free-segment straightening and rotation. Our method employs an analytical transition model similar to \cite{lui2013tangled}, but we additionally incorporate an approximate prediction of cable bending.

The third challenge is manipulation action selection. Viswanath et al. learn a manipulation policy using simulated cable states from the Blender simulation engine \cite{viswanath2021singledense}. Saha et al. generate a topological plan to change the knot configuration using a probabilistic road map \cite{saha2006dloplanning}. In \cite{nair2017combining}, monocular images of humans manipulating DLO from initial to goal state are used to guide the robot's action selection. In \cite{lui2013tangled}, optimal actions are selected by minimizing the immediate cost, but it is subject to global suboptimality, especially when a poorly-selected action leads to a state where no solution can be found. Our approach extends \cite{lui2013tangled} with a high-level action primitive selection step, which considers the number of crossings and cable distribution before performing action parameter optimization.

Finally, executing a given cable manipulation action requires consideration of the gripper hardware. Prior works have utilized different types of grippers depending on the manipulation task. In \cite{viswanath2021singledense}, a da Vinci surgical robot with miniature and precise grippers is used to manipulate thin cables with a diameter of 5mm. In \cite{viswanath2022long}, since cable tracing was considered as a manipulation primitive, a specialized parallel jaw gripper was used to cage the cables during tracing. To estimate the pose and friction of a cable, a vision-based tactile sensor was integrated into the gripper design in \cite{she2020rss}. Last but not least, \cite{buzzatto2022robotic} combines dexterous fingertips, suction cups, and Nitinol ``fingernails'' to grasp flat cables.  We demonstrate unweaving of 10 AWG electric wires and shoelaces with a diameter of 5mm using a standard 1/2-inch-wide parallel gripper on a Frank Emika.

\section{Methodology}
Our multi-cable unweaving method iteratively executes 1) a perception pipeline to construct a cable-state representation from vision feedback, 2) a valid-action identification step, and 3) a high- and low-level action selection process to eliminate all cable crossings.
\subsection{Cable State Representation and Identification} 
We represent the state of $n$ cables as a set of directed graphs with no loops or branching, denoted as $\mathcal{S} = \{G^{1}, \dots, G^{n}\}, \quad G^{i} = (\mathcal{V}^{i}, \mathcal{E}^{i})$, where $\mathcal{V}^{i}$ denotes the nodes and $\mathcal{E}^{i}$ denotes the edges. There are three types of nodes: 
\begin{itemize}
    \item Endpoints $\mathcal{V}_e$
    \item Crossings $\mathcal{V}_x$
    \item Regular nodes $\mathcal{V}_r$, where $\mathcal{V}_r = \mathcal{V} - (\mathcal{V}_e \cup \mathcal{V}_x)$
\end{itemize}

We use these directed graphs to represent the positional relationship between perceived cables. Each directed cable graph $G^{i}$ points from the free endpoint $v^{i}_{free}$ to the fixed endpoint $v^{i}_{fix}$, where $v^{i}_{free}, v^{i}_{fix} \in \mathcal{V}^{i}_e$, as shown in Fig. \ref{fig:graph}b. Each $v \in \mathcal{V}_e \cup \mathcal{V}_r$ is assigned a unique ID. The same crossings have a shared ID across different cables. We further classify $v \in \mathcal{V}_x$ into overcrossings $v_{x+}$ and undercrossings $v_{x-}$. A $v_{x+}$ on one cable is always a $v_{x-}$ on another, and vice versa. The crossing type is implicitly stored through edge labeling, where the edges connected to $v^{i}_{x+}$ are labeled with $\mathcal{E}_{+}$, the ones connected to $v^{i}_{x-}$ are labeled with $\mathcal{E}_{-}$, and the ones connected only to $v^{i}_{r}$ are labeled with $\mathcal{E}_{o}$. We ensure that an edge cannot connect to $v_{x+}$ and $v_{x-}$ at the same time.

Each node stores a 2D pixel location. To find the coordinates of each node, we first segment the RGB image into binary masks for each visually distinct cable using different color thresholds. We then discretize the cable through 2D visual tracing inspired by \cite{viswanath2022long} but using a sliding-window-based approach instead of pixel-by-pixel tracing. The window width $d_w$ is selected such that 1) the cable is approximately straight and smooth inside and in a small region in front of and behind the window and 2) the occluded part of the cable is small relative to the window width. Starting from $v^{i}_{fix}$, we slide the window towards $v^{i}_{free}$ with fixed-size steps. The sliding direction is determined by the local tangent of the cable path, computed using the binary mask of the window. As we slide the window, we compute the mean of the cable mask pixel coordinates in each window and assign this mean to the corresponding node. 
\begin{figure}[h!]
    \centering
    \includegraphics[width=0.9\linewidth]{figs/cable_graph.png}
    \caption{Cable state identification. a) Perceived RGB image of interwoven red and yellow cables, b) generated directed graphs showing the cable topology and geometry, and c) close-up view of one crossing from the cable graphs displaying node numbers and edge types, where the '+', '-', and 'o' signs on the edges denote $\mathcal{E}_{+}$, $\mathcal{E}_{-}$, and $\mathcal{E}_{o}$, respectively.}
    \label{fig:graph}
\end{figure}

To determine the type of each node in each window, we check if there are two segments separated by another cable. First, we perform a flood fill and extract and count the islands. If two islands are detected, this is a $v_{x-}$. If one island is detected, this could be 1) a $v \in \mathcal{V}_r$, if the cable path intersects the window’s bounding rectangle on at least two edges, 2) a $v_{e} \in \mathcal{V}_e$, or 3) a $v_{x-}$ located on the edge of the window. To distinguish between $v_e$ and $v_{x-}$, we further slide the window by $d_w / 2$ where $d_w$ = 35. If one or zero islands are detected in the new window, it is a $v_e$. Otherwise, it is a $v_{x-}$.

After determining the coordinates and the type for the nodes in all cables, we refine the node data by finding the overcrossings. For each $v^i_{x-}$, we find its nearest neighbor $v^j \in \mathcal{V}_r^j$, where $j \neq i$. This node is then set to an overcrossing on $G^j$ and assigned the same node ID and coordinates as the $v^i_{x-}$.

Note that we require that all crossings are physically located at least $\sqrt{2} \, d_w$ from each other and that no more than two cables form a single crossing. Given an RGB image of the cables, our graph representation fully defines the cable state and encodes both the topological relationship between the cables and the geometric information of each cable.

\subsection{Action Definition and State Transition Model}
Now we define the actions the robot will take to unweave the cables, as well as a state transition model that predicts the next state $\mathcal{S}'$, given the current state $\mathcal{S}$ and action. 

Our action vector $a = [g,\theta]$ consists of a grasp node $g \in \mathcal{V}_r$ and pivot angle $\theta \in [\theta_{min}, \theta_{max}]$. The pivot angle defines the angle that the robot will move the grasp node, following a circular path about a pivot node, denoted as $c$. The pivot node is the predecessor node of either the first $v_{x-}$ or $v_{fix}$, whichever comes first, starting from $v_{free}$. We define $L_{grasp}$ as the cable segment between $c$ and $g$ and $L_{tail}$ as the cable segment between $c$ and $v_{free}$, with lengths $l_{grasp}$ and $l_{tail}$, respectively. We define the lift height $h$ as the height at which the grasp node is lifted. Finally, we define the place point $p \in \mathbb{R}^2$ as the point where the grasp node will be placed on the table, such that $||\overrightarrow{cp}|| = l_{grasp}$ and the angle between $\overrightarrow{cp}$ and $\overrightarrow{cg}$ is $\theta$.
The steps of a single action are summarized in Algorithm \ref{algo:b}.
\vspace{-5pt}
\begin{algorithm}
Grasp the grasp node $g$\\
Lift to height $h$, where $h$ is computed using \eqref{eq:h}\\
Pull $L_{grasp}$ straight along $\overrightarrow{cg}$\\
Rotate the straightened $L_{grasp}$ by an angle $\theta$ around the pivot node $c$.
Place the cable down to the table so that $g$ is now at the place point $p$.
\caption{EXECUTE\_ACTION}
\label{algo:b}
\end{algorithm} 
\vspace{-5pt}

The key idea of our simplified and deterministic state transition model is to partially straighten the graph by each action. We assume the segment between $v_{fix}$ and $c$ stays unaltered. For $L_{grasp}$, Step 3 of Algorithm  \ref{algo:b} ensures that it is physically straightened. The state of $L_{tail}$ after the action, denoted as $L'_{tail}$, is determined by two possible state transition functions:
\begin{align} \label{eq:model}
    L'_{tail} = \left\{\begin{array}{l}
     f_{straight}(\mathcal{S}, a) \,\,\, \text{if} \,\,\, l_{tail} < k\,l_{grasp}\\[8pt]
     f_{bent}(\mathcal{S}, a)  \,\,\,\,\,\,\,\,\,\,\, \text{if} \,\,\,\, l_{tail} \geq k\,l_{grasp}
\end{array}\right.
\end{align}
where $k$ is an empirically-tuned, stiffness-dependent threshold, which we set to 0.8 in our case. If $l_{tail}<k\,l_{grasp}$, $f_{straight}$ is used, which makes $L_{tail}$ colinear with $L_{grasp}$ after the move. Otherwise, $f_{bent}$ is used, which makes $L_{tail}$ a straight segment with the line to which $L_{tail}$ belongs passing through the original $v_{free}$. This is to simulate the bending effect of the cable after we apply a normal force upwards to lift the cable and move it to a different location. Note that we assume that each action only affects one cable's graph.

\subsection{Identification of Valid Action Subspaces}
With the state transition model, we can simulate the next state and find valid actions by searching in the action space. We define the 2D workspace as the area on the table that 1) is visible by the RGBD sensor pointing downward towards the table, and 2) does not exceed the reachable workspace of the manipulator. An action is valid if it meets the following criteria:
\begin{itemize}
  \item The grasp node $g$  is at least $d_{f}$ away from any nodes in other cables, where $d_{f}$ is the gripper's fingerpad width.
  \item For the a cable's new graph $G'$ coming from a the original graph $G$, no regular node lies outside the workspace unless the cable endpoint also lies outside the workspace. This prevents a middle portion of the cable from being outside the workspace (i.e. ``broken in the middle'') and therefore causes the sliding-window-based cable tracing to fail.
  \item No nodes in $G'$ should be within a distance of $\sqrt{2} d_w$ to any $v \in \mathcal{V}_e \cup \mathcal{V}_x$ on other cables' graphs. 
\end{itemize}

We group the valid actions into subspaces in the form of \eqref{eq:sub}, where each action subspace, denoted as $\bar{A}_{sub}(g)$, is a set of actions in a continuous domain of $\theta$ and with the same predicted number of crossings eliminated, $M \in \mathbb{Z}$.
\begin{align}\label{eq:sub}
\begin{split}
    \bar{A}_{sub}(g) &= \{[a(g, \theta), M] \,\, | \,\, \theta \in [\theta_1, \theta_2]\}, \\
    \mathcal{A}_{sub} &= \{\bar{A}_{sub}(g) \, | \, g \in \mathcal{V}_r\}
\end{split} 
\end{align}

\subsection{Primitives Selection and Action Parameter Optimization}
Using the valid action subspaces, we plan the action by first selecting a high-level action primitive and then optimizing the action parameter for the selected action primitive. We define two action primitives:
\begin{enumerate}
    \item \textbf{Elimination}, which attempts to eliminate at least one crossing. 
    \item \textbf{Redistribution}, which rearranges the cables so that they are more evenly distributed, possibly without eliminating any crossings.
\end{enumerate}

To select one action primitive from these two primitives, we first iterate through every $G \in \mathcal{S}$ and check if there exists any valid action subspace $\bar{A}_{sub}(g) \in \mathcal{A}_{sub}$ with $M > 0$. If so, there exist valid elimination actions, and the elimination primitive is selected. Otherwise, no valid action subspace will result in a non-zero number of crossings eliminated, and the redistribution primitive is selected. 

With the selected action primitive, we then determine the low-level action parameters $g$ and $\theta$ by minimizing the immediate cost through gradient descent. For each primitive, a distinct cost function is used.

\begin{figure}[h!]
    \centering
    \includegraphics[width=0.85\linewidth]{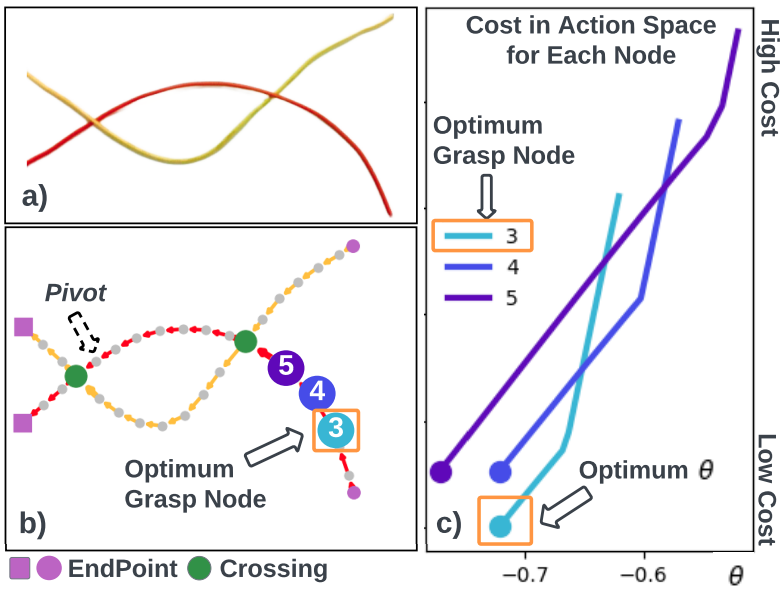}
    \caption{Elimination action optimization process. a) Image captured by the wrist camera, b) cable graphs showing the pivot node and the optimized grasp node, c) calculated cost at each graspable node in the current state, the optimum action parameter is then selected based on the lowest cost.}
    \label{fig:cost}
\end{figure}

\uline{\textit{1) Elimination Action Optimization}} aims to maximize the average distance between $G'$ to all other graphs, where the distance metric for two graphs is defined as the average L2 distance between each pair of nodes in the two graphs; minimize the new curvature at $p$; maximize the grasp-to-tail length ratio, $l_{grasp} \, / \,l_{tail}$; and 4) maximize the predicted number of eliminated crossings, $M$. The cost function is the negative of the reward defined in \eqref{eq1}, where $|\cdot|$ denotes the cardinality of a set. The optimization is performed in the union of all $\bar{A}_{sub}(g) \in \mathcal{A}_{sub}$ with $M > 0$. An example of elimination action optimization is shown in Fig. \ref{fig:cost}.
\begin{equation} \label{eq1}
\begin{split}
&r_{Elimination}(\mathcal{S}, \mathcal{S}', c, g, p) \\
= \,\, &w_{dist}\, \sum_{H \in \mathcal{S}-G} \,\, \sum_{v_i \in G', v_j \in H} ||v_i - v_j||^2_2 \, \frac{1}{|G'| |H|}\\
    + \, &w_{curv} \, Angle(\overrightarrow{c \, Succ(c)}, \overrightarrow{c \, p})\\
    + \, &w_{cred} \, \frac{l_{grasp}}{l_{tail}}\\
    + \, &w_{elim} \, M
\end{split}
\end{equation}
\vspace{-12 pt}

\uline{\textit{2) Redistribution Action Optimization}} uses a cost function similar to the one in elimination, but there are two differences: it does not take into account the number of crossings eliminated, and it additionally minimizes the standard deviation of the nodes' distance to the boundary of the workspace. For a node $v_i \in G$, we use its normalized $y$ coordinate in the image as the distance $d_i$. The cost function is the negative of the reward defined in \eqref{eq2}. It is performed in the union of all $\bar{A}_{sub}(g) \in \mathcal{A}_{sub}$ with $M = 0$.
\begin{equation} \label{eq2}
\begin{split}
&r_{Redistribution}(\mathcal{S}, \mathcal{S}', c, g) \\
= \,\, &w_{std}\, \sqrt{\frac{\sum_{v_i \in G'} (d_i - \mu)^2}{|G'|}} \\
    + \, &w_{curv} \, Angle(\overrightarrow{c \, Succ(c)}, \overrightarrow{c \, p})\\
    + \, &w_{cred} \, \frac{l_{grasp}}{l_{tail}}
\end{split}
\end{equation}

\subsection{Action Execution}
We retrieve the pixel coordinates of $g$ from the graph and look up the corresponding 3D coordinate from the depth map generated by the calibrated RGBD sensor. To execute the action, the grasp orientation is determined by the local tangent of the cable at the $g$, while the place orientation is selected to be the direction pointing from $p$ to $c$.

Another consideration is the lift height $h$. After deprojecting the cable graph nodes into 3D coordinates in the camera frame and retrieving the physical cable segment lengths, we compute the lift height $h$ using the Pythagorean theorem, where $\pi^{-1}(\cdot)$ is the deprojection operator.
\begin{align}\label{eq:h}
    h^2 = ||\pi^{-1}(c)- \pi^{-1}(g))||_2^2 - ||\pi^{-1}(c)-\pi^{-1}(p))^2||^2_2
\end{align}

Whenever a grasping action is executed, we evaluate whether we successfully grasped the cable by measuring the width $d_g$ between the gripper fingers after closing the gripper. If $d_g$ is less than half of the cable width, we consider it a grasp failure and perform a re-grasp by lowering the gripper by a small offset incrementally until grasp success. We attempt re-grasping up to five times.

\section{EXPERIMENTAL RESULTS}
We evaluate our multi-cable unweaving method through experiments conducted using different cable configurations and materials. In our experiments, we use five different types of cable configurations, including two cables with two to three crossings and three cables with three to five crossings. We quantify the success rate and the computation time of cable state identification and unweaving. 

\subsection{Cable State Identification}
To test the cable discretization and graph-building pipeline, we collect a total of 300 RGB images (640 x 480) of electric cables in 5 different cable number and crossing configurations, each with 60 images of different cable states, using the RealSense camera under indoor lighting conditions. The process is repeated 5 times for each image to ensure the consistency of the result. Cable state identification is considered successful if the graph accurately reflects the occlusion relationship and the overall shape, according to human observation. Failure occurs when an error in the cable state reconstruction, e.g. a segment of a cable is not detected, or a crossing node is classified as a regular node, resulting in a cable state representation that will break the action selection pipeline. In other cases, there are minor misclassifications that still allow for valid actions to be selected. Table \ref{table:identification} shows both the failure rate and the minor misclassification rate across all 300 images. Failures occur in less than 1\% of images, and minor misclassifications occur in less than 5\% of images. 

\begin{table}[h!]
\caption{Cable State Identification Result}
\label{table:identification}
\centering
\begin{tabular}{|c|c|c|c|c|}
\hline
\thead{Number \\of Cables} & \thead{Number of \\Crossings} & \thead{Avg. \\Processing \\Time (sec)}& \thead{Minor \\ Misclass.\\Rate (\%)} & \thead{Failure\\Rate (\%)}\\
\hline
2 & 2 & 0.092  & 1.0 & 0.0\\
\hline
2 & 3 & 0.094  & 2.6 & 0.3\\
\hline
3 & 3 & 0.122  & 3.3 & 0.0\\
\hline
3 & 4 & 0.381  & 2.6 & 0.3\\
\hline
3 & 5 & 0.405 & 4.6 & 0.7\\
\hline
\end{tabular}
\end{table}

\subsection{Cable Unweaving}
We conduct unweaving experiments with electric cables and shoelaces on the same five different types of cable configurations. For each combination, we perform ten trials with different initial cable configurations. The results are summarized in Tables \ref{table:electrical} and \ref{table:shoelaces}. 

In simple configurations with 2 cables and 2 crossings, our method has a 100\% success rate. As the number of crossings increases, the success rate is reduced. Across all trials, the average success rate is 84\%.  Although we are using a greedy planner, due to the additional redistribution primitive, we are still able to achieve 65\% success rate in the most challenging configuration tested (3 cables with 5 crossings). These results prove that our method is able to generalize across cables with different thicknesses, elasticity, and weights. 

For experiments on both types of cables, we manually tuned the scalar parameters $w_{dist}$, $w_{curv}$, $w_{cred}$, $w_{std}$, and $w_{elim}$ to be 1.0, 100, 100, 3000, and 30 respectively.

\begin{table}[h!]
\caption{Result of Robot Unweaving Electric Cables}
\label{table:electrical}
\centering
\begin{tabular}{|c|c|c|c|}
\hline
\thead{Number \\of Cables} & \thead{Number of \\Crossings} & \thead{Avg. Planning Time\\ per Action (s)} & \thead{Success\\ Rate (\%)}\\
\hline
2 & 2  &1.701 & 100\\
\hline
2 & 3 & 2.397 & 90\\
\hline
3 &3 & 2.546 & 80\\
\hline
3 & 4 & 2.954 & 70\\
\hline
3 & 5 & 3.628 & 60\\
\hline
\end{tabular}
\end{table}
\vspace*{-0.2cm}
\begin{table}[h!]
\caption{Result of Robot Unweaving Shoelaces}
\label{table:shoelaces}
\centering
\begin{tabular}{|c|c|c|c|}
\hline
\thead{Number \\of Cables} & \thead{Number of \\Crossings} & \thead{Avg. Planning Time\\ per Action (s)} & \thead{Success\\ Rate (\%)}\\
\hline
2 & 2 & 1.683 & 100\\
\hline
2 & 3 & 2.340 & 100\\
\hline
3 &3 & 2.462 & 90\\
\hline
3 & 4 & 2.889 & 80\\
\hline
3 & 5 & 3.531 & 70\\
\hline
\end{tabular}
\end{table}

 Most failure cases stem from inaccuracies in our deterministic cable state transition model, when the predicted new cable state $\mathcal{S}'$ does not match the actual new cable state after action execution. This usually happens when a small elastic deformation of the cables introduces multiple crossings in a clustered area or limited space between the crossings and the end node in the new cable state. In this scenario, our cable state identification might fail due to violations of the assumptions of our model, leading to cascading failure in the action planning process. In the case that the perception pipeline does not fail, if the space between cables after action execution is smaller than our gripper finger width $d_f$, our action space generator will deem that the gripper fingers do not have enough space to grasp the cables and thus fails to find a valid action. By comparing the results in Table \ref{table:electrical} and \ref{table:shoelaces}, Our average unweaving success rate for shoelaces is 8\% higher than the electric cables. We believe this is because the electric cables are more elastic than the shoelaces, and the elasticity is not modeled by our state transition model. 
 \begin{figure}[thpb]
      \centering
      \includegraphics[width=\linewidth]{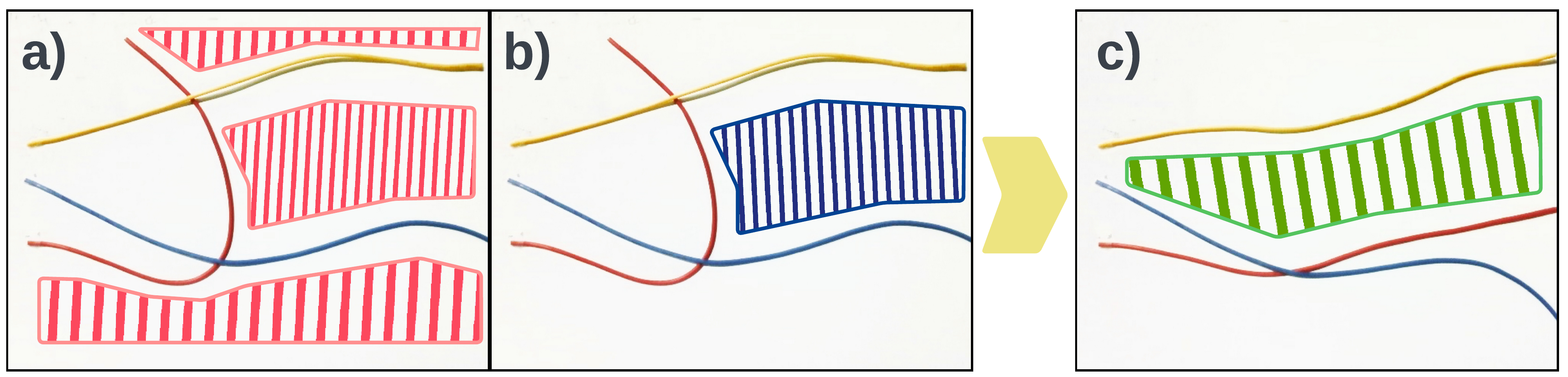}
      \caption{Action space visualization. a) No valid elimination actions exist in the red area. b) There exist valid redistribution actions in the blue area. c) After applying redistribution to the red cable in (b), a valid elimination action can then be found in the green area for the next unweaving iteration.}
      \label{fig:failure}
   \end{figure}
\subsection{Ablation Study on Action Primitives}
We also compared the performance of our approach with and without the cable redistribution step, using the electric cables. With the elimination primitive only, the unweaving success rate is shown in Table \ref{table:comp}. We observe that the redistribution primitive is crucial for finding a successful action to unweave the cable, especially as the number of crossings or the number of cables increases. Fig. \ref{fig:failure}a shows a case in which the algorithm will fail if redistribution is disabled. In this case, the elimination primitive deems that the blue cable is the only manipulable cable. A crossing can be eliminated only if the robot grasps and lifts the blue cable and pivots it about a point close to the crossing. However, according to the transition model, there is no pivot angle $\theta$ or grasp node $g$ that would result in a reduction of the number of crossings. On the contrary, when redistribution is enabled, the algorithm will attempt to pivot the red cable so that it becomes more parallel to the horizontal direction (as shown in Fig. \ref{fig:failure}b), thus creating space for pivoting the blue cable in the elimination mode.
\begin{table}[h!]
\caption{Unweaving Success Rate with the Elimination Primitive Only on Electrical Cables}
\label{table:comp}
\vspace*{-0.2cm}
\begin{center}
\begin{tabular}{|c|c|c|}
\hline
\thead{Number \\of Cables} & \thead{Number \\of Crossings} & \thead{Success \\Rate (\%)}\\
\hline
2&2 & 90 \\
\hline
2&3 & 50 \\
\hline
3&3 & 20\\
\hline
3&4 & 0 \\
\hline
3&5 & 0 \\
\hline
\end{tabular}
\end{center}
\end{table}
\vspace{-10pt}
\section{Conclusions and Future Work}
In this paper, we proposed an approach to perform autonomous multi-cable unweaving. At its core, an image-based graphical cable state representation is used to encode both the topological and the geometric information of multiple cables. We propose a state transition model for DLO(s)that generates valid action space and selects the optimal action to iteratively unweave all cables.

Our multi-cable unweaving method, which purely uses good old fashioned engineering, can serve as a baseline for future methods aiming to solve the same task. We believe that our framework is a starting point for more advanced trajectory generation and self-supervised data collection for learning-based methods because our approach can be used to collect training datasets that focus on reasonable actions rather than random and task-irrelevant actions. In terms of cable state identification, currently our method relies on different cable colors to simplify cable segmentation and crossing type identification. However, this assumption could be eliminated by learning-based methods like \cite{viswanath2022long}. Also, the state transition model could be improved by incorporating uncertainties based on cable properties determined through active perception. The relationship between cable properties and state transition model could be learned through extensive simulation or real cable data. In terms of action planning, it would be interesting to extend our greedy planner to look more steps ahead and evaluate whether non-myopic planning would improve the success rate or efficiency of the unweaving procedure. Another avenue for future work is to use learned high-level features to for the optimization instead of hand-crafted features like distance, curvature, etc.

\addtolength{\textheight}{-12cm}   



\bibliography{ref}

@inproceedings{she2020rss,
    title={Cable Manipulation with a Tactile-Reactive Gripper},
   author    = {Yu She and
               Shaoxiong Wang and
               Siyuan Dong and
               Neha Sunil and
               Alberto Rodriguez and
               Edward H. Adelson},
    booktitle={Robotics: Science and Systems (RSS)},
    year={2020}
}

@INPROCEEDINGS{nair2017combining,  
    author={Nair, Ashvin and Chen, Dian and Agrawal, Pulkit and Isola, Phillip and Abbeel, Pieter and Malik, Jitendra and Levine, Sergey},  
    booktitle={2017 IEEE International Conference on Robotics and Automation (ICRA)},  
    title={Combining self-supervised learning and imitation for vision-based rope manipulation},   
    year={2017},  
    volume={},  
    number={},  
    pages={2146-2153},  
    doi={10.1109/ICRA.2017.7989247}
}

@ARTICLE{wang2022offlineOnline,  
    author={Wang, Changhao and Zhang, Yuyou and Zhang, Xiang and Wu, Zheng and Zhu, Xinghao and Jin, Shiyu and Tang, Te and Tomizuka, Masayoshi},  
    journal={IEEE Robotics and Automation Letters},   
    title={Offline-Online Learning of Deformation Model for Cable Manipulation With Graph Neural Networks},   
    year={2022},  
    volume={7},  
    number={2}, 
    pages={5544-5551}, 
    doi={10.1109/LRA.2022.3158376}
}

@INPROCEEDINGS{aprilTag,  
author={Yang, Yuxuan and Stork, Johannes A. and Stoyanov, Todor},  
booktitle={2021 IEEE International Conference on Robotics and Automation (ICRA)},   
title={Learning to Propagate Interaction Effects for Modeling Deformable Linear Objects Dynamics},   
year={2021},  volume={},  number={},  pages={1950-1957},  
doi={10.1109/ICRA48506.2021.9561636}}

@INPROCEEDINGS{saha2006dloplanning,  
    author={Saha, M. and Isto, P.},  
    booktitle={Proceedings 2006 IEEE International Conference on Robotics and Automation, 2006. ICRA 2006.},   
    title={Motion planning for robotic manipulation of deformable linear objects},   
    year={2006},  volume={},  
    number={},  
    pages={2478-2484},  
    doi={10.1109/ROBOT.2006.1642074}
}

@INPROCEEDINGS{schulman2013tracking,  
    author={Schulman, John and Lee, Alex and Ho, Jonathan and Abbeel, Pieter},  
    booktitle={2013 IEEE International Conference on Robotics and Automation},   
    title={Tracking deformable objects with point clouds},   
    year={2013},  
    volume={},  
    number={},  
    pages={1130-1137},  
    doi={10.1109/ICRA.2013.6630714}
}

@INPROCEEDINGS{lui2013tangled,  
    author={Lui, Wen Hao and Saxena, Ashutosh},  
    booktitle={2013 IEEE/RSJ International Conference on Intelligent Robots and Systems},   
    title={Tangled: Learning to untangle ropes with RGB-D perception},   
    year={2013},  
    volume={},  
    number={},  
    pages={837-844},  
    doi={10.1109/IROS.2013.6696448}
}

@article{wireHarness,
title = {Manufacturing automation for automotive wiring harnesses},
journal = {Procedia CIRP},
volume = {97},
pages = {379-384},
year = {2021},
note = {8th CIRP Conference of Assembly Technology and Systems},
issn = {2212-8271},
author = {Huong Giang Nguyen and Marlene Kuhn and Jörg Franke}}

@INPROCEEDINGS{industry,
  TITLE = {{Robotic Manipulation and Sensing of Deformable Objects in Domestic and Industrial Applications: A Survey}},
  AUTHOR = {Sanchez, Jose and Corrales Ramon, Juan Antonio and Bouzgarrou, Belhassen-Chedli and Mezouar, Youcef},
  JOURNAL = {{The International Journal of Robotics Research}},
  PUBLISHER = {{SAGE Publications}},
  VOLUME = {37},
  NUMBER = {7},
  PAGES = {688 - 716},
  YEAR = {2018},
  MONTH = Jun,
}

@inproceedings{viswanath2021singledense,
    author = {Viswanath, Vainavi and Grannen, Jennifer and Sundaresan, Priya and Thananjeyan, Brijen and Balakrishna, Ashwin and Novoseller, Ellen and Ichnowski, Jeffrey and Laskey, Michael and Gonzalez, Joseph E. and Goldberg, Ken},
    title = {Disentangling Dense Multi-Cable Knots},
    year = {2021},
    publisher = {IEEE Press}
}

@inproceedings{viswanath2022long,
    AUTHOR    = {Vainavi Viswanath AND Kaushik Shivakumar AND Justin Kerr AND Brijen Thananjeyan AND Ellen Novoseller AND Jeffrey Ichnowski AND Alejandro Escontrela AND Michael Laskey AND {Joseph E.} Gonzalez AND Ken Goldberg}, 
    TITLE     = {Autonomously Untangling Long Cables}, 
    BOOKTITLE = {Proceedings of Robotics: Science and Systems}, 
    YEAR      = {2022}, 
    DOI       = {10.15607/RSS.2022.XVIII.034} 
}

@inbook{manturov2020knottheory, 
    place={Boca Raton}, 
    title={Knot theory}, 
    publisher={CRC Press}, 
    author={Manturov, V. O.}, 
    year={2020}, 
    pages={4–10}
}

@inproceedings{buzzatto2022robotic,
  title={On Robotic Manipulation of Flexible Flat Cables: Employing a Multi-Modal Gripper with Dexterous Tips, Active Nails, and a Reconfigurable Suction Cup Module},
  author={Buzzatto, Joao and Chapman, Jayden and Shahmohammadi, Mojtaba and Sanches, Felipe and Nejati, Mahla and Matsunaga, Saori and Haraguchi, Rintaro and Mariyama, Toshisada and MacDonald, Bruce and Liarokapis, Minas},
  booktitle={2022 IEEE/RSJ International Conference on Intelligent Robots and Systems (IROS)},
  pages={1602--1608},
  year={2022},
  organization={IEEE}
}

\end{document}